\documentclass[sigconf]{acmart}

\AtBeginDocument{%
  }

\begin{document}
\raggedbottom
\title{MALM: Mask Augmentation based Local Matching for Food-Recipe Retrieval}

\author{Bhanu Prakash Voutharoja}
\affiliation{%
 \institution{University of Wollongong}
 \streetaddress{Northfields Ave}
 \city{Wollongong}
 \state{New South Wales}
 \country{Australia}}
\email{voutharoja.bhanu06@gmail.com}

\author{Peng Wang}
\affiliation{%
 \institution{University of Wollongong}
 \streetaddress{Northfields Ave}
 \city{Wollongong}
 \state{New South Wales}
 \country{Australia}}
\email{pengw@uow.edu.au}

\author{Lei Wang}
\affiliation{%
 \institution{University of Wollongong}
 \streetaddress{Northfields Ave}
 \city{Wollongong}
 \state{New South Wales}
 \country{Australia}}
\email{leiw@uow.edu.au}

\author{Vivienne Guan}
\affiliation{%
 \institution{University of Wollongong}
 \streetaddress{Northfields Ave}
 \city{Wollongong}
 \state{New South Wales}
 \country{Australia}}
\email{vguan@uow.edu.au}


\begin{abstract}
Image-to-recipe retrieval is a challenging vision-to-language task of significant practical value. The main challenge of the task lies in the ultra-high redundancy in the long recipe and the large variation reflected in both food item combination and food item appearance.  A de-facto idea to address this task is to learn a shared feature embedding space in which a food image is aligned better to its paired recipe than other recipes. However, such supervised global matching is prone to supervision collapse, i.e., only partial information that is necessary for distinguishing training pairs can be identified, while other information that is potentially useful in generalization could be lost. To mitigate such a problem, we propose a mask-augmentation-based local matching network (MALM), where an image-text matching module and a masked self-distillation module benefit each other mutually to learn generalizable cross-modality representations.  On one hand, we perform local matching between the tokenized representations of image and text to locate fine-grained cross-modality correspondence explicitly. We involve representations of masked image patches in this process to alleviate overfitting resulting from local matching especially when some food items are underrepresented. On the other hand, predicting the hidden representations of the masked patches through self-distillation helps to learn general-purpose image representations that are expected to generalize better. And the multi-task nature of the model enables the representations of masked patches to be text-aware and thus facilitates the lost information reconstruction. Experimental results on Recipe1M dataset show our method can clearly outperform state-of-the-art (SOTA) methods. Our code will be available at XXXXX.
\end{abstract}




\maketitle

\section{Introduction}
\label{sec:intro}
\begin{figure}[t]
    \begin{center}
    \includegraphics[width=1.0 \linewidth]{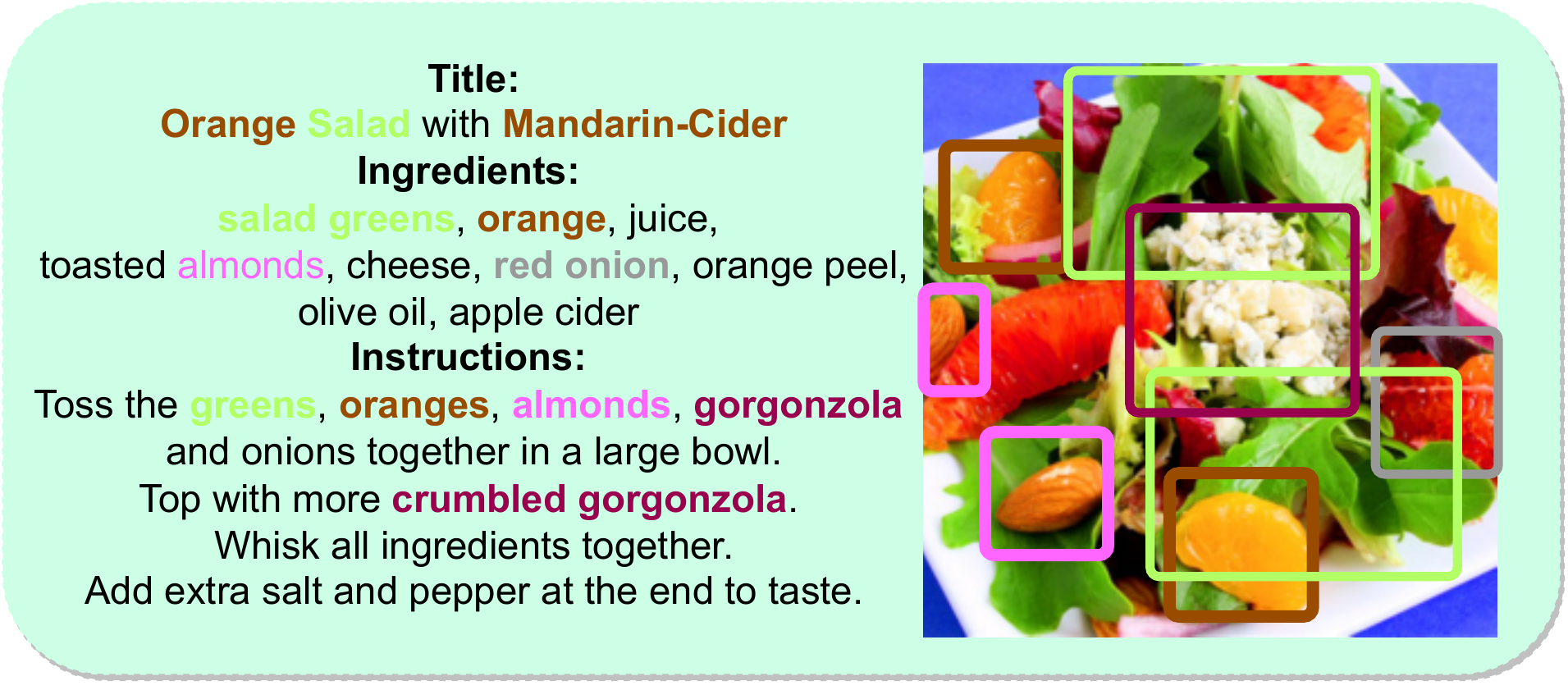}
    \caption{Illustration of the complexity of recipe and food images in the image-to-recipe retrieval task. The correspondence between the food image and recipe is shown via bounding boxes in the color of the highlighted text.}
    \label{fig:introduction_figure}
    \end{center}
\end{figure}
Food consumption is closely linked to our health and cultures~\cite{recipe1m+}. How to use computer vision techniques to advance this fundamental human experience has significant practical value. Image-to-recipe retrieval is one of such vision task that observes wide applications, such as digital cooking, dietary tracking, and food recommendation, just to name a few. This task has attracted great research attention since the release of Recipe1M~\cite{Salvador} and Recipe1M+~\cite{recipe1m+}, large-scale, structured corpus of over one million cooking recipes and 13 million food images.

Image-to-recipe retrieval is a challenging task compared to standard image-text retrieval. Fig. \ref{fig:introduction_figure} illustrates the complexity of the image-recipe retrieval task. Firstly, the textual recipes are normally quite lengthy, and much content in the recipes is irrelevant to the retrieval task. For example, in Recipe1M, the recipe consists of three entities, i.e., title, ingredients, and instructions, and each instruction contains 208 words on average~\cite{recipe1m+}. Secondly, the image tends to observe a wide range of food item combinations and the food items contained normally have fine-grained nature and large intra-class variation. To address this challenging task, most existing works~\cite{revamping, TFood, MCEN, HF-ICMA} focused on designing effective encoders to extract useful features from both modalities such that in the shared feature space paired image-recipes are close while unpaired ones are pushed apart. However, such supervised global-level image and recipe representation matching may suffer from supervision collapse~\cite{crosstransformer}, i.e., only partial information that is necessary for distinguishing training pairs can be identified while other useful information that is desirable for generalization will be lost. 

To alleviate the supervision collapse problem, in this work, we propose a mask-augmentation based local matching (MALM) network. Our model adopts a multi-task training strategy, where an image-text matching module aligns paired images and recipes, and a masked self-distillation module learns general-purpose image representations. Importantly, these two modules complement each other and can work together to develop cross-modality representations that can generalize better. \textbf{Firstly}, thanks to unified tokenized representations from Transformer~\cite{transformer}, we propose a local-matching based contrastive loss that matches the image patch representations against the local text features to explicitly learn fine-grained correspondence. However, instead of performing local matching on top of all the raw image patch representations, the majority of the image patches are masked out and the masked representations are involved in the image-text matching. This can be regarded as data augmentation, which can effectively alleviate the potential overfitting resulting from local matching especially when some food items are underrepresented. \textbf{Secondly}, the hidden representations of the masked image patches are predicted based on a self-distillation module. That is, the model first adopts a teacher model to produce the representations of the original image patches which are then reconstructed by a student model based on a masked version of the input. The parameters of the teacher model are updated as an exponentially moving average of the student network. Note that different from existing design~\cite{datatovec}, the masked representations in our model are enforced to be text-aware through the image-text matching module and thus can facilitate the lost information reconstruction for missing patches. This is expected to be able to alleviate the difficulty of visual representation reconstruction under extremely complex food data variation.  
The experiments on Recipe1M dataset show the appealing performance of the proposed MALM model.


The contributions of this work can be summarized as follows:
\begin{itemize}
\item We approach the challenging image-to-recipe retrieval task from the perspective of supervision collapse and propose a token-based local matching strategy to address the supervision collapse by explicitly locating the fine-grained cross-modality correspondence. 

\item We propose a novel multi-task model, where a text-aware masked self-distillation is proposed to alleviate the overfitting resulted from local matching and consequently learn general-purpose visual features that generalize better.

\item Experimental results on Recipe1M show the proposed method can achieve new state-of-the-art image-to-recipe retrieval performance and reveal the appealing properties of the proposal in addressing supervision collapse. 

\end{itemize}

\section{Related Work}
\label{sec:related_work}

Since the release of food datasets like Food-101~\cite{Food-101} and ISIA Food-500~\cite{ISIA-Food-500}, the computer vision community has made tremendous strides in the field of food detection. The majority of works concentrate on classifying food images~\cite{deepfood, saki, food-recognition, neutrinet, chinesefoodnet, cleanet}, with the objective being to establish the category of the food image. Other studies investigate a variety of tasks, including calculating the number of ingredients in a dish~\cite{food-quantity-estimation, pita}, determining calories~\cite{Im2calories}, and guessing contents using multiple labels~\cite{deep-based-ingredients, food-attributes}. Since the publication of multi-modal datasets like Recipe1M~\cite{Salvador} and Recipe1M+~\cite{recipe1m+}, new tasks involving the use of both visual and written recipes have evolved. Several studies put forth solutions for cross-modal recipe retrieval~\cite{ACME, deep-understanding, Salvador, Adamine, revamping, TFood}, recipe generation~\cite{structure-aware, pita, procedural, procedural-photo, inversecooking}, and question answering~\cite{recipeqa} that make use of image-recipe paired data. 

\subsection{Vision-Language Representation Learning}
In recent years, vision-language research has advanced rapidly. A number of different cross-modality loss functions have been proposed for the training objective, including image-text matching~\cite{vilbert, uniter}, masked language modeling~\cite{bert}, masked image modeling~\cite{beit, Flava}, and contrastive learning~\cite{infonce, simclr}. These are often combined to create a compound objective. Few works based on contrastive learning techniques~\cite{MOCO, simclr, SimSiam, BYOL, DINO} specifically investigate the effectiveness of learning visual representations for image classification. Multi-modal (image and text) contrastive learning objectives~\cite{clip, align, SLIP} have recently achieved promising performance in learning strong visual representations. Multi-modal contrastive learning extends contrastive learning to two modalities, such as images and text. This approach has been shown to be effective in learning strong visual representations. For example, the CLIP model~\cite{clip} uses a multi-modal contrastive learning objective to learn visual representations. The multi-modal contrastive learning objectives~\cite{clip, align, SLIP} have recently achieved promising performance in learning strong visual representations. These approaches have the potential to revolutionize the way we train vision-language models.

\subsection{Cross-Modal Retrieval}
Finding the right sample in one modality given a data sample in a different modality, and vice versa is the goal of the cross-modal retrieval task. The cosine-similarity score is calculated using the embeddings of data samples of both modalities in a common space, and the sample with the highest score is then retrieved. For this task, methods often involve text and image encoding with LSTM or Transformer text encoders and pre-trained deep convolutional neural networks~\cite{TFood}. For recipe encoding, initial approaches use word2vec~\cite{word2vec} and skip-thoughts~\cite{skip-vectors} to embed the words and sentences, which are then encoded using recurrent networks (e.g., LSTMs). For better alignment of image and text, \cite{Messina} uses a transformer encoder for image-sentence alignment. Additionally, cross-modal retrieval has benefited from the application of adversarial learning~\cite{ACME, R2GAN, chefgan}. Some studies have demonstrated the advantages of using attention to capture the intricate connections between visual and language, which improves the joint embedding space~\cite{stacked, itm}. In order to improve regional-level and regional-global linkages, Wen et al. \cite{LearningDS} execute graph attention. Through cross-modal message aggregations, it has been successfully demonstrated that the interaction of multi-modal data can be strengthened~\cite{Sain, CAMP}. 
The alignment of the ingredients and the instructions are not equal; some ingredients are plainly visible in the image while others are not. This has led to some research into adding attention modules to recipe encoders or image encoders to weigh various tokens and regions differently when fusing the two modalities~\cite{MCEN, HF-ICMA, SCAN}. As a result of the success of transformers in text and vision, some recent work has been done to utilize transformers, with encouraging results.

Cooking recipes, in contrast to the brief descriptions from captioning datasets, are lengthy, structured text documents that are difficult to encode~\cite{revamping}. Chen et al.~\cite{deep-understanding} use hierarchical attention to model each recipe component independently for strong recipe feature extraction. R2GAN~\cite{R2GAN} use an adversarial technique, to enhance the learning of recipe features by creating images from recipes. Another work, ACME~\cite{ACME} uses an adversarial learning technique coupled with a retrieval learning sample strategy for effective cross-modal alignment. Moreover, \cite{SN} encodes three components of the recipe separately using three attention networks and improves the triplet loss to lessen the impact of noise by optimizing the most extreme hard negative sample. Recently, \cite{revamping} proposed a hierarchical transformer-based encoder specifically for recipes and achieved SOTA performance. The hierarchical transformer consists of three transformer encoders one for each recipe component to extract sentence-level embeddings. Next, another transformer aligns these embeddings to achieve intra-fusion using a self-supervised loss. Finally, the recipe embedding is obtained by concatenation of aligned titles, instructions, and ingredient sentence-level embeddings. Another work~\cite{cooking-programs}, generates cooking programs conditioned on the image and recipes. For each recipe, they generate a set of valid cooking program sequences. In the inference stage, the trained model not only retrieves the image/recipe but also predicts the cooking program. Shukor et al.~\cite{TFood} proposed a framework called TFood, which has image and recipe encoders with additional multi-modal regularization and image-text matching blocks to promote cross-alignment between image and text features. They further propose an adaptive triplet loss with a dynamic margin that adjusts the hardness of the learning process. We use TFood~\cite{TFood} as our baseline because of its higher performance, but we swap out their naive image-text matching with our suggested method of two-level image-text matching with mask-based data augmentation.


\section{Proposed Method}
\label{sec:methodology}

\begin{figure*}
    \begin{center}
    \includegraphics[width=1.0 \textwidth]{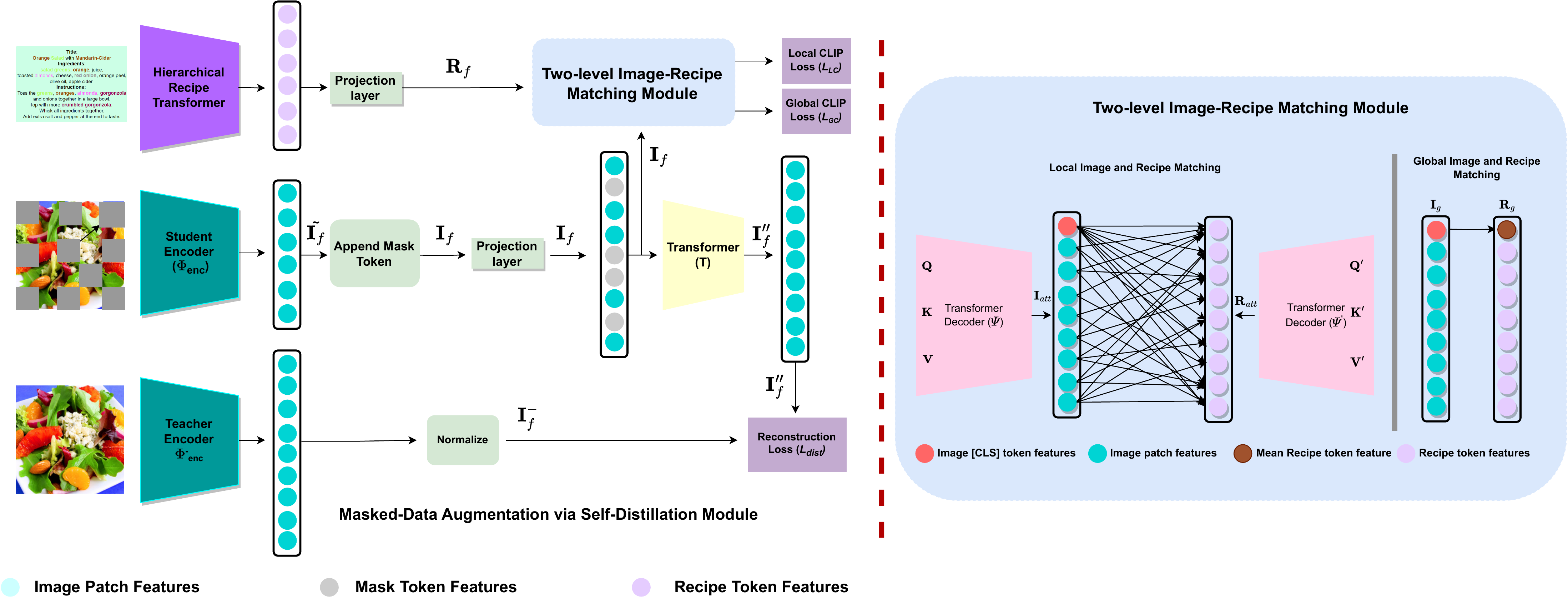}
    \caption{Illustration of our MALM framework. Our proposed framework has two modules - a) Two-level Image Recipe Matching to learn fine-grained image and recipe features at both local and global levels and thus alleviates supervision collapse; b) Masked-Data Augmentation via Self-Distillation for learning more generalized image features. Due to the multi-task nature of our model, the image representations learned by the student encoder are text-aware since we first perform the image-recipe matching on both masked and visible tokens and later use these matched features for masked image feature reconstruction.}
    \label{fig:malm}
    \end{center}
\end{figure*}

Fig.~\ref{fig:malm} shows the overall framework of the proposed MALM model. Given a set of food images and their corresponding recipes, we pass the images to an image encoder and recipes to a text encoder to extract the visual and textual features respectively. We use vision transformer (ViT)~\cite{ViT} initialized with CLIP (Contrastive Language-Image Pre-Training)~\cite{clip} pretrained weights as the image encoder. Each recipe has three components: title, instructions, and ingredients.  Similar to TFood~\cite{TFood}, we use a hierarchical recipe encoder that has three transformer encoders for extracting sentence-level features of each recipe component and a transformer decoder with self and cross-attention to capture interactions between all the three recipe components for a strong intra-fusion. The final output of the hierarchical recipe encoder is a feature vector which is a concatenation of cross-attention features of the title, ingredients, and instructions. After extracting the image and recipe features, we explicitly perform image and recipe matching at local and global levels. Detailed discussion regarding the image-recipe matching is available in Sec \ref{sec:image_text_matching}. The involvement of masked image features in image-recipe matching leads to more generalized and recipe-aware reconstructed image features. Hence, we regard this as masking-based data augmentation which is detailed in Sec \ref{sec:data_augmentation}. \\
\indent The global-level image-recipe matching may not sufficiently consider the finer details of the image, which could lead to supervision collapse On the other hand, the proposed two-level image-recipe matching approach effectively addresses this issue. By learning strong features from both global and local levels, the model can alleviate supervision collapse. In Section \ref{sec:supervision_collapse}, the effectiveness of the proposed approach, referred to as MALM, is demonstrated by comparing it with \cite{TFood}.


\subsection{Data Augmentation using Mask-based Self-Distillation}
\label{sec:data_augmentation}

For each image, we mask out a large subset of the image patches and replace the representations of masked patches with a mask token for image-text matching, which can be regarded as a data-augmentation to avoid overfitting.  Inspired by the previous work~\cite{datatovec, maskclip}, we use a masked self-distillation strategy to reconstruct the masked image patch features. Specifically, the model first adopts a teacher model to produce the representations of the original image patches which are then reconstructed by a student model based on a masked version of the input. The parameters of the teacher model are updated as an exponential moving average of the student network. But our model differs from this typically masked self-distillation from two perspectives. Firstly, the masked self-distillation in~\cite{datatovec} is proposed purely as a loss for self-supervised pre-training, while in our model it is employed as a data-augmentation operation to avoid overfitting from cross-modality matching. Secondly, the student network in our module interacts with the text through the multi-task nature of our network and thus makes the masked representation from the student to be text-aware. And such context information will be important to facilitate the missing information reconstruction especially when dealing with complex food images with large variations.

Let $\phi_{enc}$ and $\phi^{-}_{enc}$ represent the student and teacher image encoders respectively. Given an input image $\textit{I}$, we pass it through $\phi^{-}_{enc}$ to extract the features $\phi^{-}_{enc}(I) = \mathbf{I}^{-}_{f} = \{\mathbf{f}_{cls},\mathbf{f}_{1},...,\mathbf{f}_{p}\}$. At the same time, we randomly mask some patches of $\textit{I}$, and feed the unmasked patches $\tilde{\textit{I}}$ to $\phi_{enc}$. Let $\mathcal{M}$ be the set of indices of masked patches. The features extracted by $\phi_{enc}$ are $\mathbf{\tilde{I}}_{f} = \{\mathbf{\tilde{f}}_{cls}\} \cup \{\mathbf{\tilde{f}}_{p \notin \mathcal{M}}\}$. The masked features are replaced with a learnable mask token denoted as $m$ to form a complete set of features $\mathbf{I}_{f} = \{\mathbf{\tilde{f}}_{cls},\mathbf{\tilde{f}}_{1},...,\mathbf{\tilde{f}}_{p}\}$, with $\mathbf{\tilde{f}}_{i \in \mathcal{M}} = m$. The mask-appended features are then projected and passed to an image-recipe matching module that performs matching at both local and global levels. The same features are also fed into a single-layer Transformer encoder ($\textit{T}$) to predict the features for the missing patches,
\begin{equation}
    \textit{T}(\mathbf{I}_{f}) = \mathbf{I}^{\prime\prime}_{f} = \{\mathbf{f}^{\prime\prime}_{cls},\mathbf{f}^{\prime\prime}_{1},...,\mathbf{f}^{\prime\prime}_{p}\}.
\end{equation}

To match the target features generated by $\phi^{-}_{enc}$ i.e. $\mathbf{I}^{-}_{f}$, with the predicted features $\mathbf{I}^{\prime\prime}_{f}$, we use a distillation loss,
\begin{equation}
    \mathcal{L}_{dist} = \dfrac{1}{|{\mathcal{M}}|}\sum_{p \in \mathcal{M}} \text{SmoothL1}(\mathbf{f}^{\prime\prime}_{p},\text{StopGradient}(\mathbf{f}_{p}),\beta),
\end{equation}
where $L1$ loss with a smoothing factor $\beta$ is employed as the loss function. StopGradient() prevents the gradient update of teacher image encoder ($\phi^-_{enc}$)

The local matching and masked distillation modules are used only during the training phase. For inference, the output features from image and recipe encoders are directly used for retrieval. 

\subsection{Image-Recipe Matching}
\label{sec:image_text_matching}


To align the image and recipe features in the embedding space, we do image-text matching using a contrastive loss~\cite{clip} at both local-level and global-level. 
The local matching is motivated by the observation that global-representation based matching can risk losing local information within both food image and recipes, thus deteriorating the representation capacity of the features. 

The right side of Fig \ref{fig:malm} shows our proposed two-level image-text matching module. Let $\mathbf{I}_{f}$ be the masked image features from student encoder $\phi_{enc}$ and $\mathbf{R}_{f}$ be the recipe features extracted by a recipe encoder. We pass these extracted image and recipe features through two different projection layers to match their dimensionality. Next, the projected image features are passed through a single layer transformer decoder ($\psi_{dec}$) with image features ($\mathbf{I}_{f}$) as input and recipe feature $\mathbf{R}_{f}$ as context. First, the queries, keys, and values are calculated via linear transformation of image and recipe features i.e., queries $\mathbf{Q} = \mathbf{W}_{q}.\mathbf{I}_{f}$, keys $\mathbf{K} = \mathbf{W}_{k}.\mathbf{R}_{f}$ and values $\mathbf{V} = \mathbf{W}_{v}.\mathbf{R}_{f}$ where $\mathbf{W}_{q}$, $\mathbf{W}_{k}$, and $\mathbf{W}_{v}$ are trainable weights used to perform linear transformation. 
The cross-attention between image and recipe features is calculated as a scaled dot-product of queries and keys, i.e., 

\begin{equation}
    \mathbf{A}_{I} = softmax(\dfrac{\mathbf{Q}\mathbf{K}^T}{\sqrt{D}}),
\end{equation}
\begin{equation}
    \mathbf{I}_{att} = A_{I}\mathbf{V},
\end{equation}
where $A_{I} \in \mathbf{R}^{B \times P \times S}$ represents the attention weights for recipe features obtained using a softmax function $\text{softmax}()$. $\mathbf{I}_{att} \in \mathbf{R}^{B \times P \times D}$ represents the cross-attention image features. Here $B$ is batch size, $P$ is the number of patches in the image, $S$ is the recipe sequence length, and $D$ is the feature dimensions.

We follow the same method to obtain cross-attention recipe features using another single-layer transformer decoder ($\psi_{dec}^\prime$) with queries $\mathbf{Q}^{\prime} = \mathbf{{W}_{q}}^{\prime}.\mathbf{R}_{f}$, keys $\mathbf{K}^{\prime} = \mathbf{{W}_{k}}^{\prime}.\mathbf{I}_{att}$ and values $\mathbf{V}^{\prime} = \mathbf{{W}_{v}}^{\prime}.\mathbf{I}_{att}$ where $\mathbf{{W}_{q}}^{\prime}$, $\mathbf{{W}_{k}}^{\prime}$, and $\mathbf{{W}_{v}}^{\prime}$ are trainable weights used to perform linear transformation. The cross-attention recipe features are obtained as 

\begin{equation}
    \mathbf{A}_{R} = softmax(\dfrac{\mathbf{Q}^{\prime}\mathbf{{K}^{\prime}}^T}{\sqrt{D}}),
\end{equation}
\begin{equation}
    \mathbf{R}_{att} = \mathbf{A}_{R}\mathbf{V}^{\prime},
\end{equation}
where $\mathbf{A}_{R} \in \mathbf{R}^{B \times S \times P}$ represents the attention weights obtained using a softmax function $\text{softmax()}$, and $\mathbf{R}_{att} \in \mathbf{R}^{B \times S \times D}$ represents the cross-attention recipe features.

Once the cross-attention image and recipe features are obtained, the image-recipe matching is done at global and local levels. To do the global-level matching, we extract the global image and recipe features. The additional $[\textit{CLS}]$ token used in vision transformer~\cite{ViT} is considered a global token since it attends to all patches of the image. Hence, features of $[\textit{CLS}]$ are considered as global features of the image i.e., $\mathbf{I}_{g} = \mathbf{I}_{att}[:,CLS,:]$. The global recipe features are obtained by taking the average of cross-attention recipe features i.e., $\mathbf{R}_{g} = \dfrac{\sum_{s=1}^{S} \mathbf{R}_{att}[:,s,:]}{S}$.
The global image-recipe matching is performed using contrastive loss, an objective function that promotes semantically similar representations for the paired data and contrastive representations for unpaired data. In a batch of B samples, there would be B positive pairs and $B^2-B$ negative pairs. The positive pairs are pulled close to each other while the negative pairs are pushed apart. Let $Z$ be a contrastive function as

\begin{equation}
    Z(\mathbf{x}_{i},\mathbf{y}_{i}) = \dfrac{\exp(\mathbf{x}_{i} \cdot \mathbf{y}_{i} / \tau)}{\sum_{\substack{k=1 , k \neq i}}^{B} \exp(\mathbf{x}_{i} \cdot \mathbf{y}_{k} / \tau)},
\end{equation}
where $\mathbf{x}_{i}$, and $\mathbf{y}_{i}$ are the features to be matched and B is the batch size.
The CLIP loss performs both the image-text and text-image matching and returns its average as the final loss. Since we use CLIP loss for image-recipe matching, our global-level clip loss can be obtained as 

\begin{equation}
    \mathcal{L}_{GC} = \sum_{i = 1}^{B} \dfrac{Z(\mathbf{I}_{g}[i,:], \mathbf{R}_{g}[i,:]) + Z(\mathbf{R}_{g}[i,:], \mathbf{I}_{g}[i,:])}{2}.
\end{equation}

To learn cross-modality representations that can explicitly reflect the fine-grained correspondence, we propose to use a local-level image-recipe matching, which aligns the patch-wise image features with its relevant recipe features. The relevant recipe features for each image patch are obtained by performing elementwise multiplication of patch-wise softmax attention weights with cross-attention recipe features, i.e,
\begin{equation}
\label{eq:9}
    \mathbf{R}_{{l}_{p}} = \mathbf{A}_{I}[:,p,:] \odot \mathbf{R}_{att}
\end{equation}

where $p = \{1,2,...,P\}$ represents number of patches in the image and $\mathbf{R}_{l_{p}}$ represents weighted recipe features relevant to each image patch. We then compute mean of $\mathbf{R}_{{l}_{p}}$ as 
$\mathbf{R}_{{l}_{p}} = \dfrac{\sum_{s = 1}^{S} \mathbf{R}_{{l}_{p}}[:,s,:]}{S}$. The local image-recipe matching is performed as 
\begin{equation}
    \mathcal{L}_{LC} = \sum_{i=1}^{B} \dfrac{\sum_{p=1}^{P} (Z(\mathbf{I}_{att}[i,p,:],\mathbf{R}_{l_{p_{i}}}) + Z(\mathbf{R}_{l_{p_{i}}},\mathbf{I}_{att}[i,p,:]))}{P}.
\end{equation}

\subsection{Training Objective}

We use TFood~\cite{TFood} without the image-text matching module as our baseline model. We replace their naive image-text matching module with our proposed two-level (local and global) matching module coupled with masked self-distillation. Our final training objective is
\begin{equation}
    \mathcal{L} = \mathcal{L}_{itc} + \lambda_{itm}(\mathcal{L}_{GC} + \mathcal{L}_{LC}) + \lambda_{dist}{\mathcal{L}_{dist}},
\end{equation}
where $\mathcal{L}_{itc}$ is the semantic triplet loss from TFood, $\lambda_{itm}$ and $\lambda_{dist}$ are the weights for image-text matching and reconstruction losses respectively. The input to $\mathcal{L}_{itc}$ are image features from student encoder and recipe features.

\section{Experiments}
\label{sec:results}
This section shows the experimental results to verify the effectiveness of our suggested strategy, including comparisons to existing solutions and ablations studies to reveal appealing properties of the proposed method.

\begin{table*}[!ht]
  \begin{center}
  \scalebox{0.90}{
    \begin{tabular}{c|cccc|cccc|cccc|cccc}
     \specialrule{0em}{1pt}{1pt}
     \toprule[1.2pt]
     & \multicolumn{8}{c|}{{\textbf{1K}}} & \multicolumn{8}{c}{{\textbf{10K}}} \\ 
     \specialrule{0em}{1pt}{1pt}
     \cline{2-17}
     \specialrule{0em}{1pt}{1pt}
        & \multicolumn{4}{c|}{{\textbf{image-to-recipe}}} & \multicolumn{4}{c|}{{\textbf{recipe-to-image}}} &
        \multicolumn{4}{c|}{{\textbf{image-to-recipe}}} &
        \multicolumn{4}{c}{{\textbf{recipe-to-image}}} \\ 
        \specialrule{0em}{1pt}{1pt}
     \cline{2-17}
     \specialrule{0em}{1pt}{1pt}
     ~ & medR & R-1 & R-5 & R-10 & medR & R-1 & R-5 & R-10 & medR & R-1 & R-5 & R-10 & medR & R-1 & R-5 & R-10 \\ 
     \specialrule{0em}{1pt}{1pt}
     \hline
     \specialrule{0em}{1pt}{1pt}
     Salvador et al.~\cite{Salvador} & 5.2 & 24.0 & 51.0 & 65.0 & 5.1 & 25.0 & 52.0 & 65.0 & 41.9 & - & - & - & 39.2 & - & - & - \\ 
     Adamine~\cite{Adamine} & 2.0 & 40.2 & 68.1 & 78.7 & 2.0 & 39.8 & 69 & 77.4 & 13.2 & 14.8 & 34.6 & 46.1 & 14.2 & 14.9 & 35.3 & 45.2 \\ 
     R2GAN~\cite{R2GAN} & 2.0 & 39.1 & 71.0 & 81.7 & 2.0 & 40.6 & 72.6 & 83.3 & 13.9 & 13.5 & 33.5 & 44.9 & 12.6 & 14.2 & 35.0 & 46.8 \\ 
     MCEN~\cite{MCEN} & 2.0 & 48.2 & 75.8 & 83.6 & 1.9 & 48.4 & 76.1 & 83.7 & 7.2 & 20.3 & 43.3 & 54.4 & 6.6 & 21.4 & 44.3 & 55.2 \\ 
     ACME~\cite{ACME} & 1.0 & 51.8 & 80.2 & 87.5 & 1.0 & 52.8 & 80.2 & 87.6 & 6.7 & 22.9 & 46.8 & 57.9 & 6.0 & 24.4 & 47.9 & 59.0 \\ 
     SN~\cite{SN} & 1.0 & 52.7 & 81.7 & 88.9 & 1.0 & 54.1 & 81.8 & 88.9 & 7.0 & 22.1 & 45.9 & 56.9 & 7.0 & 23.4 & 47.3 & 57.9 \\ 
     IMHF~\cite{IMHF} & 1.0 & 53.2 & 80.7 & 87.6 & 1.0 & 54.1 & 82.4 & 88.2 & 6.2 & 23.4 & 48.2 & 58.4 & 5.8 & 24.9 & 48.3 & 59.4 \\ 
     Wang et. al~\cite{wang} & 1.0 & 53.5 & 81.5 & 88.8 & 1.0 & 55.0 & 82.0 & 88.8 & 6.0 & 23.4 & 48.8 & 60.1 & 5.6 & 24.6 & 50.0 & 61.0 \\ 
     SCAN~\cite{SCAN} & 1.0 & 54.0 & 81.7 & 88.8 & 1.0 & 54.9 & 81.9 & 89.0 & 5.9 & 23.7 & 49.3 & 60.6 & 5.1 & 25.3 & 50.6 & 61.6 \\ 
     HF-ICMA~\cite{HF-ICMA} & 1.0 & 55.1 & 86.7 & 92.4 & 1.0 & 56.8 & 87.5 & 93 & 5.0 & 24.0 & 51.6 & 65.4 & 4.2 & 25.6 & 54.8 & 67.3 \\ 
     MSJE~\cite{MSJE} & 1.0 & 56.5 & 84.7 & 90.9 & 1.0 & 56.2 & 84.9 & 91.1 & 5.0 & 25.6 & 52.1 & 63.8 & 5.0 & 26.2 & 52.5 & 64.1 \\ 
     SEJE~\cite{SEJE} & 1.0 & 58.1 & 85.8 & 92.2 & 1.0 & 58.5 & 86.2 & 92.3 & 4.2 & 26.9 & 54.0 & 65.6 & 4.0 & 27.2 & 54.4 & 66.1 \\ 
     M-SIA~\cite{M-SIA} & 1.0 & 59.3 & 86.3 & 92.6 & 1.0 & 59.8 & 86.7 & 92.8 & 4.0 & 29.2 & 55.0 & 66.2 & 4.0 & 30.3 & 55.6 & 66.5 \\ 
     DaC~\cite{DaC} & 1.0 & 60.2 & 84.0 & 89.7 & 1.0 & - & - & - & 4.0 & 30.0 & 56.5 & 67.0 & - & - & - & - \\
     X-MRS~\cite{X-MRS} & 1.0 & 64.0 & 88.3 & 92.6 & 1.0 & 63.9 & 87.6 & 92.6 & 3.0 & 32.9 & 60.6 & 71.2 & 3.0 & 33.0 & 60.4 & 70.7 \\ 
     H-T~\cite{revamping} & 1.0 & 60.0 & 87.6 & 92.9 & 1.0 & 60.3 & 87.6 & 93.2 & 4.0 & 27.9 & 56.4 & 68.1 & 4.0 & 28.3 & 56.5 & 68.1 \\ 
     H-T (ViT)~\cite{revamping} & 1.0 & 64.2 & 89.1 & 93.4 & 1.0 & 64.5 & 89.3 & 93.8 & 3.0 & 33.5 & 62.1 & 72.8 & 3.0 & 33.7 & 62.2 & 72.7 \\ 
     T-Food (CLIP-ViT)~\cite{TFood} & 1.0 & 72.3 & 90.7 & 93.4 & 1.0 & 72.6 & 90.6 & 93.4 & 2.0 & 43.4 & 70.7 & 79.7 & 2.0 & \textbf{44.6} & 71.2 & 79.7 \\
     Baseline & 1.0 & 66.2 & 85.1 & 88.9 & 1.0 & 66.2 & 85.3 & 88.5 & 2.0 & 38.3 & 65.9 & 75.4 & 2.8 & 37.0 & 65.2 & 75.5 \\
     \hline 
     \specialrule{0em}{1pt}{1pt}
     \textbf{MALM}  & \textbf{1.0} & \textbf{74.0} & \textbf{91.3} & \textbf{94.3} & \textbf{1.0} & \textbf{73.0} & \textbf{91.0} & \textbf{93.9} & \textbf{2.0} & \textbf{45.9} & \textbf{72.3} & \textbf{80.5} & \textbf{2.0} & 44.2 & \textbf{71.7} & \textbf{80.1} \\
     \hline 
    \end{tabular}
   }
  \end{center}
  \caption{\textbf{Comparison with Previous Methods.} medR ($\downarrow$), R-K ($\uparrow$) are reported on Recipe1M test set for 1k and 10k test sizes. The best score for each column is highlighted in bold. Our Baseline is T-Food (CLIP-ViT), without their image-text matching (ITM) module.}
  \label{tab:table1}
 \end{table*}

\subsection{Dataset}
We use the Recipe1M dataset~\cite{Salvador} to train, validate, and test our model. Recipe1M is a large dataset of recipes that was collected from cookery websites. The dataset contains 1,029,720 recipes, each of which includes a title, a list of ingredients, a list of preparation instructions, and optionally an image. We split the dataset into three sets: training, validation, test with 720,639, 155,036, and 154,045 recipes. We only use paired data to train, validate, and test our model. The number of paired images and recipes in the training, validation, and test sets is 238,999, 51,119, and 51,303 respectively.

\subsection{Evaluation Metrics}
Following the previous works \cite{TFood}, we employ the median rank (medR) and recall metrics R-K (with K = 1, 5, 10) to assess the effectiveness of our model. Our evaluation entails the computation of average metrics on 10 bags of 1k samples and 5 bags of 10k samples randomly chosen from test set.

\begin{table}[!ht]
    \centering
     \scalebox{0.95}{
    \begin{tabular}{|c|c|c|c|c|}
        ~ & medR & R-1 & R-5 & R-10 \\ \hline
        Baseline & 2.0 & 38.3 & 65.9 & 75.4 \\ 
        Baseline + $\mathcal{L}_{LC}$ & 2.0 & 41.7 & 68.5 & 76.2 \\ 
        Baseline + $\mathcal{L}_{LC}$ + $\mathcal{L}_{GC}$ & 2.0 & 43.8 & 70.3 & 79.5 \\ 
        Baseline + $\mathcal{L}_{LC}$ + $\mathcal{L}_{GC}$ + $\mathcal{L}_{dist}$ & \textbf{2.0} & \textbf{44.2} & \textbf{71.7} & \textbf{80.1} \\
        MALM + masking both modalities & 2.0 & 43.9 & 71.1 & 79.8 \\
    \end{tabular}
    }
    \caption{Ablation studies. medR ($\downarrow$), R-K ($\uparrow$) are reported on Recipe1M test set for image-to-recipe retrieval task with 10k test size.}
    \label{tab:table2}
\end{table}

\begin{table}[!ht]
    \centering
    \begin{tabular}{|c|c|c|c|c|}
        Mask Ratio & medR & R-1 & R-5 & R-10 \\ \hline
        0.90 & 2.0 & 43.8 & 70.1 & 79.8 \\
        0.75 & \textbf{2.0} & \textbf{44.2} & \textbf{71.7} & \textbf{80.1} \\ 
        0.50 & 2.0 & 41.7 & 68.5 & 76.2 \\ 
        0.25 & 2.0 & 39.8 & 66.3 & 75.2 \\ 
    \end{tabular}
    \caption{Comparison of image-to-recipe retrieval results for various image masking ratios on 10k test setup}
    \label{tab:table3}
\end{table}

\subsection{Implementation Details}
\label{sec:qualitative}
\begin{figure*}
    \begin{center}
    \includegraphics[width=1.0 \textwidth]{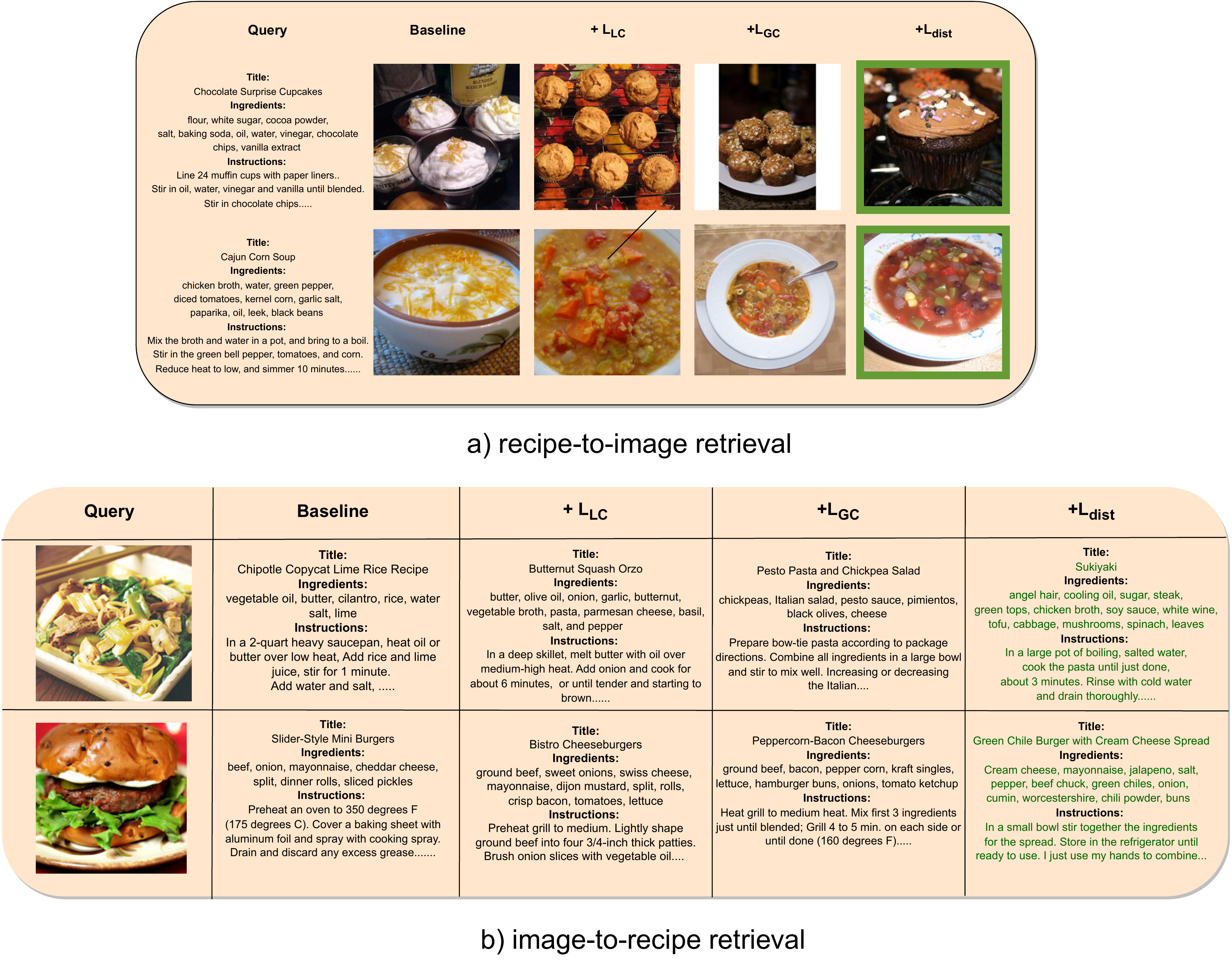}
    \caption{Qualitative Analysis. We show the top-1 retrieved image for the input recipe query. Our baseline is TFood~\cite{TFood} but without their image-text matching module. $\textbf{L}_{LC}$ refers to local-level clip loss ($\mathcal{L}_{LC}$). $\textbf{L}_{GC}$ refers to global-level clip loss ($\mathcal{L}_{GC}$). $\textbf{L}_{dist}$ refers to self-distillation reconstruction loss ($\mathcal{L}_{dist}$). The ground-truth image is highlighted with a green colour.}
    \label{fig:qualitative_analysis}
    \end{center}
\end{figure*}

Our experimental setup closely resembles that of the study by TFood~\cite{TFood}. We utilized ViT-B/16, a vision transformer, as our image encoder, which was initialized with CLIP weights, resulting in CLIP-ViT-B/16. Our hierarchical recipe transformer comprises two transformer encoders, each with 2 layers and 4 heads, with a hidden size of 512. The output dimension of the two projection layers is 768. The transformers in our two-level image-recipe matching module consist of 4 layers and 4 heads. To reconstruct the masked image patch features, we utilized a transformer encoder with 2 layers and 2 heads, with a hidden size of 768. The values of $\lambda_{dist}$ and $\lambda_{itm}$ are set to 1.0, and 1.0 respectively. The image encoder remains frozen for the first 20 epochs, after which all modules are trained using the Adam optimizer, with a learning rate of $1e-5$, except for CLIP-ViT-B/16, which is trained with a learning rate of $1e-6$. We trained the models with a batch size of 128 for 120 epochs. All transformers, except for CLIP-ViT-B/16, were trained from scratch. Our training utilized 2 NVIDIA V100 GPUs, each with 32GB of VRAM.

\subsection{Comparison with Existing Solutions}
In Table \ref{tab:table1}, we compare the results obtained from our model with those of previous works. Since our baseline is TFood~\cite{TFood}, the results of previous studies are quoted from TFood paper for fair comparison. Our proposed two-level (local and global) image-recipe matching module coupled with mask-based data augmentation could enhance the baseline results by +7.3 \%, +5.7 \%, +5.4 \% on R-1, R-5, and R-10 metrics, respectively, on the test set (10k) for image-to-recipe retrieval task thus making our model (MALM) new state-of-the-art (SOTA). On the recipe-to-image retrieval task, we also acheive compelling results with an improvement of +7.1 \%, +6.5 \%, +4.6 \% on R-1, R-5, and R-10 metrics ,respectively, on test set (10k). By replacing the image-text matching module in TFood with our proposed image-text matching, we could enhance the performance on the image-recipe retrieval task by +5.1 \%, +3.1 \%, +3.1 \% improvement in R-1, R-5, and R-10 metrics, respectively, on the test set (10K). Similarly, we could achieve consistent improvement even on the recipe-image retrieval task with an average improvement of +3.7 \% on the 10k test setup across all the recall metrics. On the 1k test setup, our proposed method achieved an average improvement of +2.5 \% for recall metrics. Moreover, when compared with other SOTA models such as H-T (ViT) \cite{revamping}, the performance gain is much more significant with an increase of +11.9 \% for R-1, +10.1 \% for R2 and +7.5 \% for R3 on image-recipe retrieval task on 10k test setup. Furthermore, the performance gap between the existing cross-attention-based methods such as HF-ICMA~\cite{HF-ICMA}, M-SIA~\cite{M-SIA} and our proposed MALM is much higher with an average improvement of +20.6 \% and +15.1 \% respectively under 1k and 10k test setup. 

\subsection{Qualitative Analysis}
Fig. \ref{fig:qualitative_analysis} shows qualitative analysis on recipe-to-image and image-to-recipe retrieval tasks. The image retrieved by the baseline model for recipe query in row 1 of Fig. \ref{fig:qualitative_analysis}a is a basic cupcake without reflecting any fine-grained ingredients such as "chocolate", "white sugar", "chocolate chips", "vanilla", or "muffin cups with paper liners". Upon adding our proposed local-level clip loss for image-recipe matching, our model was able to identify "chocolate", and "cocoa powder" and retrieved an image accordingly. Moreover, the global-level image-text matching further helped the model to identify ingredients such as "flour", "white sugar" and "vanilla". Finally, after adding the self-distillation loss for image reconstruction, our model was able to identify much more fine-grained ingredients such as "paper liners" and "24 muffin cups" and retrieved the best matching image to the recipe query. Even for the recipe query in row 2 of Fig. \ref{fig:qualitative_analysis}a, our model with image-recipe matching and self-distillation modules, was able to identify keywords such as "corn", "soup", "chicken", "tomatoes", "black beans", and "green bell pepper" in the title, ingredients, and instruction components of recipe query and retrieve the best image. The qualitative analysis of raw images and recipes shows the effectiveness of performing image-recipe matching on two levels. Moreover, coupling the image-text matching module with the self-distillation module helps the image encoder to learn more generalizable recipe-correlated image features. The same performance reflects even on the image-to-recipe retrieval task. In row 1 of Fig. \ref{fig:qualitative_analysis}b, given a food image as input, our approach with all the modules could retrieve the exact ground-truth recipe by identifying key ingredients such as "chicken broth", "spinach leaves", and "cabbage" which was missed by the baseline.

\subsection{Ablation Study}
\label{sec:ablation_study}
We conduct an ablation study starting with a baseline and adding each module one at a time, recording the increase in performance, to evaluate the significance of various modules in our model. Our baseline is TFood but without their image-text matching (ITM) module. We then add our local-level image-text clip loss $\mathcal{L}_{LC}$ which improved the scores of recall metrics by 1.3 \%. Next, by adding global-level clip loss $\mathcal{L}_{GC}$, the performance is further improved. Our image-recipe matching module with both local-level and global-level image-recipe matching could improve the baseline recall scores by an average of + 2.3 \%. Next, by adding our mask-based self-distillation loss $\mathcal{L}_{dist}$, we could further enhance the R-1 score by + 2.9 \%, R-5 score by + 1.7 \% and R-10 score by + 1.3 \%. Overall, by adding our proposed image-text matching module and data augmentation using a mask-based self-distillation module to our baseline, we could achieve SOTA performance across all the recall metrics.

\begin{figure*}[!h]
    \begin{center}
    \includegraphics[width=1.0 \textwidth]{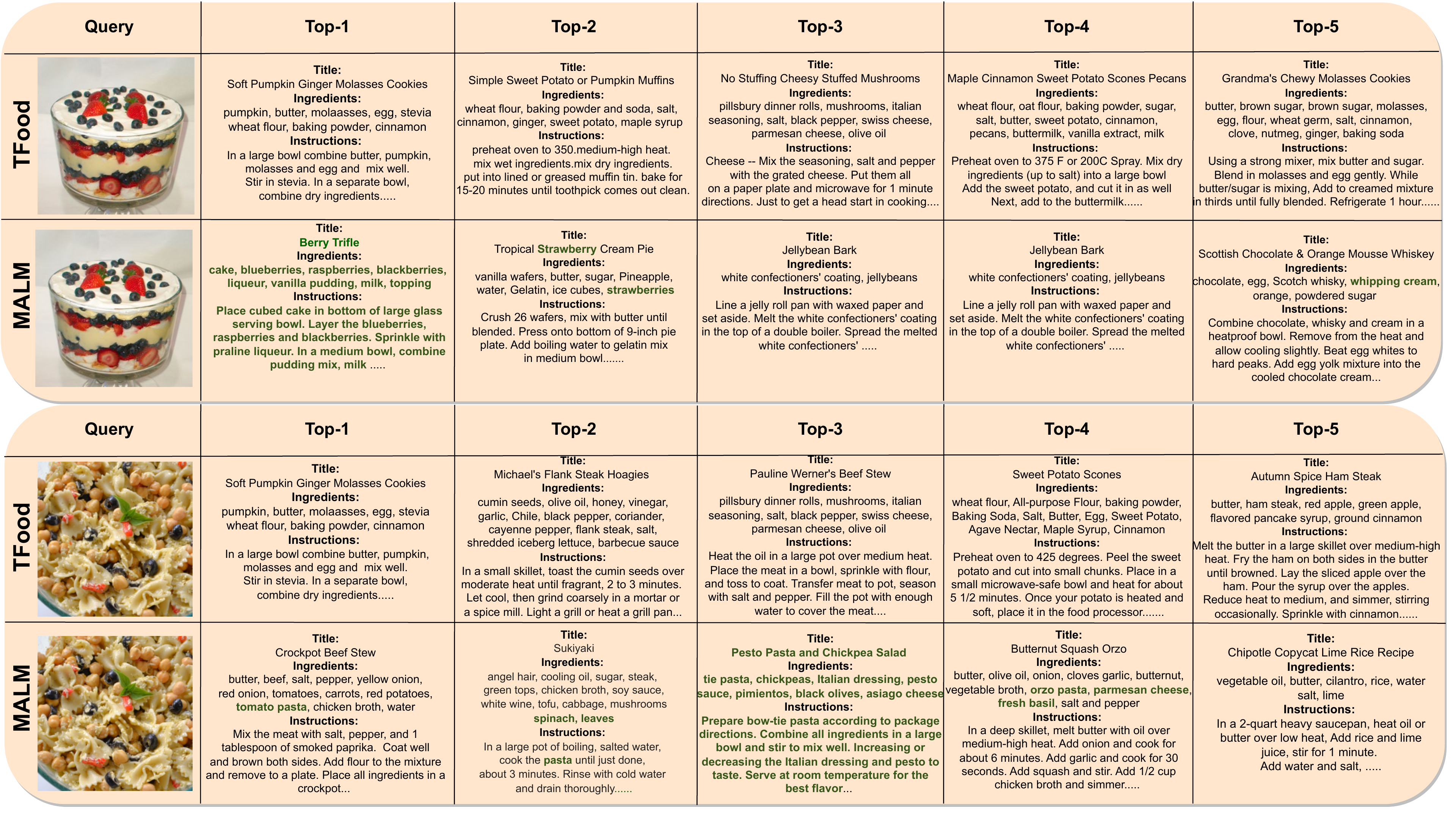}
    \caption{This figure demonstrates instances of supervision collapse observed in TFood, as well as the efficacy of our proposed model in addressing this issue. The ground-truth recipes are highlighted with green colour.}
    \label{fig:supervision_collapse}
    \end{center}
\end{figure*}

\indent To evaluate the influence of the image-patch masking ratio on the overall performance, we conduct experiments on our model with four different masking ratios - 0.90, 0.75, 0.5, and 0.25. Results are available in Table \ref{tab:table3}. When the masking ratio is 0.75, the performance gain under recall metrics is at its maximum, while when it is 0.25, the performance gain is at its lowest. By masking the majority of the image patches, the model is forced to capture the image's rich local patterns in order to reconstruct them. This operation is also expected to be able to alleviate overfitting result from dense local matching. \textbf{Masking both food images and recipes} We conducted an experiment to investigate the applicability of our approach in masking both recipe tokens and image patches, where we randomly mask them with a masking ratio of 0.75 and then reconstruct the recipes in the same manner as images. The results presented in Table \ref{tab:table3} indicate that the masking of both the image and recipe modalities may not yield a substantial advantage when compared to masking the food images alone. This could be attributed to the image modality's rich semantic information that aligns with the recipe information, allowing the model to reconstruct the food image using recipe data. Conversely, a recipe comprises three distinct components, namely title, ingredients, and instructions, with an average concatenation length of 574 tokens. It contains redundant information that is irrelevant to the food image, such as the instruction to "Preheat an oven for 10 minutes at 80 degrees". As a result, the clues extracted from the food image might be inadequate to fully reconstruct the recipe, ultimately resulting in decreased performance.

\subsection{Supervision Collapse}
\label{sec:supervision_collapse}


To investigate the issue of supervision collapse in the baseline model and the effectiveness of our proposed approach in mitigating it, we conducted an analysis and comparison of the top-5 recipes retrieved by both the TFood and MALM models. Fig. \ref{fig:supervision_collapse} depicts examples of supervision collapse. In the first instance, the TFood model identified the query image as a cookie and retrieved recipes pertaining to cookies and muffins. However, it failed to recognize certain local details in the image, such as strawberries and blackberries, and relied heavily on more general information, such as chocolate chips, cream, and sugar. Additionally, the shape of the query image resembled that of a muffin. In contrast, our proposed framework, which utilizes two-level matching and reconstructed recipe-aware image features, was able to capture important nuances such as the presence of berries, vanilla pudding, and cake, resulting in the retrieval of recipes that more closely aligned with the image. Even when a pasta image was used as the query, MALM retrieved four out of five recipes containing the term 'pasta'. Moreover, our model successfully identified fine-grained ingredients such as cheese, basil, black olives, and leaves from the image, as evidenced by the highlighted regions in green in Fig. \ref{fig:supervision_collapse}, which were also reflected in the retrieved recipes.  

\section{Conclusion}
\label{sec:conclusion}
In this work, we investigated the image-text retrieval task from the perspective of supervision collapse, that is, performing supervised global text-image matching can result in a loss of information that is not necessary for fitting training data but desirable for generalization. To address this problem, we proposed a mask-augmentation-based local matching model, which employs two important modules that can benefit each other mutually to learn cross-modality features that generalize better.  A local matching module locates fine-grained cross-modality correspondence and provides external supervision for a masked self-distillation module. The masked self-distillation module learns general-purpose image features and avoids overfitting caused by local matching.  


\begin{acks}
This research was supported by Multiple Sclerosis Australia Incubator Grant (Grant no. 21-1-110), Illawarra Health and Medical Research Institute Young Investigator Award, NCI Adapter Allocation Scheme and UOW/NCI Partner Share Scheme. VG was supported by Multiple Sclerosis Australia Postdoctoral Fellowship (Grant no. 20-223).
\end{acks}

\bibliographystyle{ACM-Reference-Format}
\bibliography{acmart}



\end{document}